\documentclass[10pt,twocolumn,letterpaper]{article}

\usepackage{iccv}
\usepackage{times}
\usepackage{epsfig}
\usepackage{graphicx}
\usepackage{amsmath}
\usepackage{amssymb}

\usepackage{adjustbox}
\usepackage{array}
\usepackage{booktabs}
\usepackage{colortbl}
\usepackage{float,wrapfig}
\usepackage{hhline}
\usepackage{multirow}
\usepackage{subcaption} 
\usepackage{enumitem}
\usepackage[accsupp]{axessibility}  

\usepackage[pagebackref=true,breaklinks=true,colorlinks,bookmarks=false]{hyperref}

\iccvfinalcopy 

\def\iccvPaperID{12517} 

\ificcvfinal\fi

\begin{document}

\title{Simultaneously Boosting Efficiency and Capability for Linear Attention via Adaptive Channel Compression and Activation Purification}

\author{
    Senqi Guan,~Xianchao Zhang,~Moyan Tian,~Linlin Zong,~Wenxin Liang,~and~Xinyue Liu \\
    Dalian University of Technology, Liaoning, China
}

\maketitle
\ificcvfinal\thispagestyle{empty}\fi

\begin{abstract}
Self-attention's quadratic complexity constitutes a core bottleneck for Vision Transformers (ViTs). Existing efficiency-oriented optimizations, including softmax-based and linear attention (LA) variants, generally compromise model representational capacity, leading to inevitable performance degradation. We propose a novel linear attention which alleviates this efficiency-capability dilemma. For the first time, we embed an Adaptive Channel Compression and Activation Purification (ACCAP) mechanism into the QKV projection of linear attention, which enables the model to retain high-dimensional feature information while performing computations with reduced channels. We further fuse a focused kernel function and a self-gating module into ACCAP to construct a lightweight linear attention module (ACCAP-LA). Our theoretical analysis shows that, compared to standard linear attention, ACCAP-LA reduces the attention module's parameter count by 79.17\% and FLOPs by 77.5\%. Extensive experiments demonstrate ACCAP-LA’s effectiveness: when applied to ViT, it achieves 79.4\% Top-1 accuracy on ImageNet-1K with only 5.8M parameters, and more notably, it boasts strong cross-architecture compatibility, boosting EfficientViT-b1 and PVT-T by 3.3\% and 3.4\% respectively. 
\end{abstract}

\section{Introduction}
Vision Transformers (ViTs) \cite{dosovitskiy2020image} have become the dominant paradigm in visual representation learning, delivering state-of-the-art(SOTA) performance on image classification \cite{DBLP:conf/icml/YaoWW0L24} and object detection \cite{DBLP:conf/icml/0008DSCT025} via global self-attention modeling. However, the quadratic complexity $(O(N^{2}))$ of vanilla softmax self-attention \cite{vaswani2017attention} emerges as a core bottleneck hampering their further scaling and deployment.
\begin{figure}[t]
\centering
    \centerline{\includegraphics[width=\columnwidth]{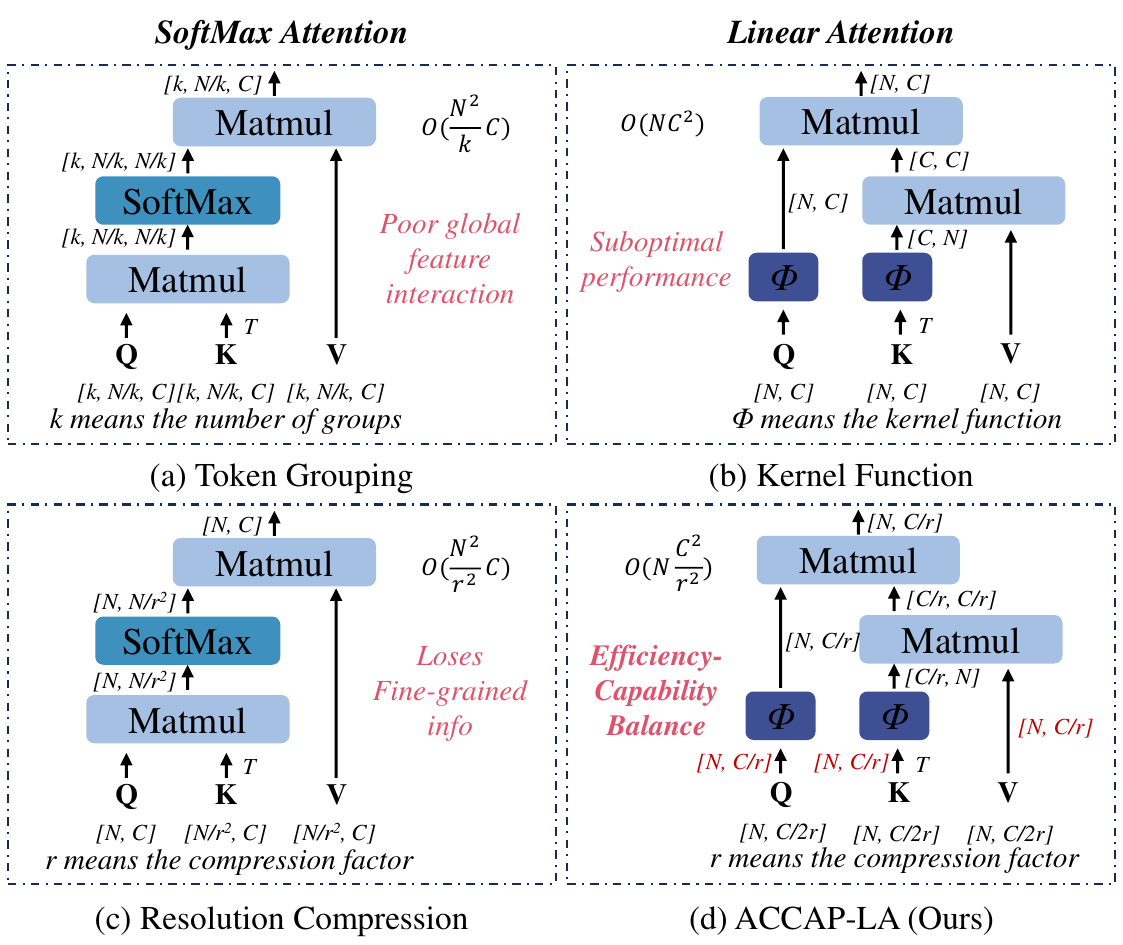}}
    \caption{
      Schematic diagram of the structural comparison between typical attention mechanisms, improved strategies, and inherent limitations (softmax attention: (a), (c); linear attention: (b), (d)).}
    \label{fig:structural_comparison}
\end{figure}
To address this issue, numerous efficiency-oriented optimizations have emerged, falling into two main categories: softmax-based approaches and linear attention-based ones. Yet all variants under these two categories compromise model capability in different aspects, leading to performance degradation.

\begin{figure*}[t]
\centering
    \centerline{\includegraphics[width=2\columnwidth]{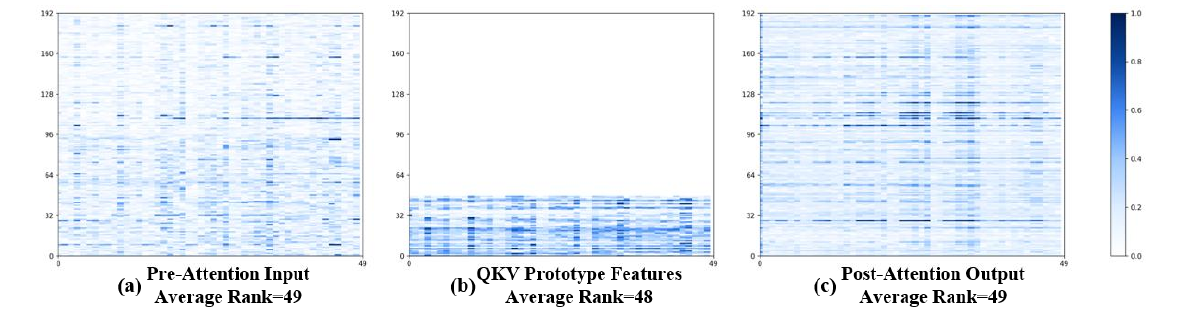}}
    \caption{
      Feature map visualization with matrix rank annotations. (a) Pre-attention input features. (b) QKV prototype features generated by the proposed compressed activation mechanism. (c) Post-attention output features.}
    \label{fig:visualization}
\end{figure*}

Optimizations for softmax attention can be split into two distinct strategies: token grouping typified by Swin Transformer \cite{liu2021swin}, and resolution compression represented by PVT v2 \cite{wang2022pvt}. However, as shown in Figure \ref{fig:structural_comparison}(a) and (c), token grouping impairs the model's capacity of global information perception, while resolution compression sacrifices its fine-grained feature sensitivity.

Linear attention \cite{katharopoulos2020transformers} emerges as a novel attention paradigm that abandons the softmax function. By leveraging the associative law of multiplication, it rearranges the computation order to first compute the Key-Value product ($K^{\mathrm{T}}V$) before multiplying with Query ($Q$), thereby reducing the complexity from quadratic to linear. Representative works based on linear attention make targeted optimizations, such as Flatten Transformer \cite{han2023flatten} which introduces a kernel function and S3ML \cite{zhang2025s3ml} which integrates sampled softmax. However, as illustrated in Figure \ref{fig:structural_comparison}(b), linear attention can essentially be regarded as a low-rank decomposition and approximation of softmax attention, so even with these improvements, its performance remains unsatisfactory. Furthermore, existing linear attention optimizations mainly focus on the computation stage, leaving channel-wise design in the core QKV projection stage largely underexplored.

To address this efficiency-capability dilemma, we propose a novel Adaptive Channel Compression and Activation Purification (ACCAP) mechanism, which we embed into the linear attention QKV projection process for the first time (Figure \ref{fig:structural_comparison}(d)). Since more channels enable richer multi-dimensional feature representations, ACCAP first allows the model to access such high-dimensional feature representations. It then enables QKV matrices to learn high-dimensional feature semantics via low-dimensional vectors during generation, with actual attention computation performed using fewer channels. Specifically, ACCAP comprises three stages: channel compression, activation purification, and channel expansion, which eliminates computational redundancy in linear attention and effectively preserves and refines critical feature information in the attention module. Intuitively, as illustrated in Figure \ref{fig:visualization}, the feature rank remains nearly stable through this pipeline, indicating that channel reduction incurs negligible information loss.

Notably, the ACCAP mechanism reduces the complexity of standard linear attention from $O(NC^2)$ to $O(NC^2/r^2)$, where $r$ denotes the compression ratio. Furthermore, we sequentially incorporate a focused kernel function and a self-gating module following ACCAP to construct the ACCAP-augmented Linear Attention (ACCAP-LA). When $r=2$, theoretical derivations confirm that compared to standard linear attention, ACCAP-LA reduces the parameter count by 79.17\% and FLOPs by 77.5\%.

Building upon ACCAP-LA, we design a series of ViT models (dubbed ACCAP-ViT-M0 to M3) with varying parameter sizes based on a unified architecture. To validate their effectiveness, we conduct image classification experiments on the ImageNet-1K dataset and object detection experiments on the COCO 2017 dataset. Across models of all scales, ACCAP-ViTs consistently deliver outstanding performance. We summarize our contributions as follows:

\begin{itemize}
    \item We propose a channel-wise optimization mechanism dubbed ACCAP for linear attention QKV projection, which reduces computational redundancy while alleviating the inherent representational degradation of vanilla linear attention.
    \item We develop a novel lightweight linear attention module ACCAP-LA by integrating ACCAP with a focused kernel function and a self-gating module, achieving a superior efficiency-performance trade-off.
    \item Extensive experiments on ImageNet-1K and COCO 2017 validate that ACCAP-LA achieves competitive performance, high inference efficiency, and strong cross-architecture generalization across various ViT backbones.
\end{itemize}

\section{Related Work}
\subsection{Efficiency Optimization of SoftMax Attention in Vision Transformers}
Existing optimizations for softmax attention in ViTs mainly follow two paradigms: token grouping and resolution compression. Token grouping methods, typified by Swin Transformer \cite{liu2021swin}, restrict attention within local windows or dynamic clusters to cut computational overhead. Resolution compression approaches, represented by PVT v2 \cite{wang2022pvt}, reduce token counts via pyramid downsampling. While effective in lowering complexity, both paradigms operate predominantly on the token dimension and inevitably impair global perception or fine-grained feature sensitivity, leaving channel-wise computational redundancy largely unexplored.

\subsection{Performance Enhancement for Linear Attention-Based Models}
Linear attention (LA) achieves linear complexity by eliminating softmax and reordering computation, yet its inherent low-rank nature causes severe representational degradation. Existing enhancements mainly revolve around three directions: kernel function design (e.g., Flatten Transformer \cite{han2023flatten} with focused kernels to sharpen attention distribution), multi-scale feature integration (e.g., EfficientViT \cite{cai2023efficientvit} to enrich semantic contexts), and hybrid architectures that combine LA with softmax or state-space designs. Nevertheless, these strategies yield limited gains or sacrifice inherent linear complexity advantages. Notably, most existing improvements focus on the attention computation stage, while the QKV projection process, which dominates LA's parameter and computation budget, remains underexplored from a channel-wise perspective.

\subsection{Channel Compression and Gated Activation Mechanisms}
The squeeze-and-excitation (SE) paradigm \cite{hu2018squeeze} pioneers channel-wise adaptive feature reweighting via compression and gating. Follow-ups like CBAM \cite{WooCBAM} and ECA-Net \cite{wangECA2020} further optimize this paradigm from spatial extension and compression efficiency perspectives. All these methods target feature quality improvement, using channel compression only for attention weight generation rather than cost reduction. SCConv \cite{li2023scconv} confirms prevalent channel redundancy in visual models, but also aims at feature refinement instead of cutting attention computation overhead. Gated mechanisms are also widely explored in attention research. Qiu et al. \cite{qiu2026gated} apply self-gating to boost non-linearity and alleviate attention sinks in LLMs. For visual linear attention, works like HLSM \cite{Guan2026HybridLA} introduce channel gating in feature fusion stages for diversity enhancement, yet none modify the core QKV projection pipeline.

Beyond attention-oriented channel designs, inverted residual bottlenecks \cite{sandler2018mobilenetv2} in convolutional networks and low-rank matrix factorization \cite{wang2020linformer} in attention modules also adopt channel-wise dimension reduction, but both are essentially pure parameter compression strategies without dedicated feature purification mechanisms. More critically, neither of them has been customized to address the inherent representational degradation issue of linear attention from the perspective of QKV projection optimization.

Despite their widespread use in feature enhancement and computation reduction, channel compression and gating have not been systematically integrated into the QKV projection of linear attention to jointly reduce computational cost and mitigate representational degradation. To this end, we propose the ACCAP mechanism.

\section{Methods}
To address these limitations mentioned above, this paper proposes a novel linear attention mechanism, named ACCAP-LA, from the perspective of feature channel redundancy \cite{li2023scconv}. Specifically, we design an Adaptive Channel Compression and Activation Purification (ACCAP) mechanism explicitly tailored for the QKV projection process, and further complement it with a focused kernel function and a self-gating module. This design ensures the model maintains lightweight properties and high efficiency, while significantly boosting its representational capability. Detailed descriptions of the proposed ACCAP-LA mechanism are presented as follows.

\subsection{Adaptive Channel Compression and Activation Purification Mechanism}
\subsubsection{Why ACCAP is Necessary?}
Feature channel redundancy is widely verified in existing vision models \cite{li2023scconv, zhou2024adapt}, which motivates the necessity of channel compression in the QKV projection process. The number of channels represents different dimensions of feature representations: a higher channel dimension enables the model to capture multi-faceted semantic information from input features, thereby enhancing the expressiveness of the attention module. However, an excessively large channel dimension introduces substantial computational and memory overhead during the QKV projection and subsequent attention computation, which contradicts the goal of lightweight model design. To reconcile the trade-off between representational capacity and computational efficiency, we propose to leverage high-dimensional channels for rich feature learning, while employing ACCAP for channel compression and activation purification before computation. This design enables the model to encode high-dimensional feature information into a low-dimensional space, achieving competitive performance with significantly reduced computational costs.

\subsubsection{Detailed Steps of ACCAP}
Let the input feature map be denoted as $X \in \mathbb{R}^{C \times H \times W}$, where $C$ is the number of input channels, and $H,W$ represent the height and width of the feature map, respectively. In vanilla linear attention, the QKV matrices are generated by directly mapping $X$ to $3C$ channels via a single $1\times1$ convolution, followed by splitting along the channel dimension:
\begin{equation}
[Q, K, V] = \text{Split}(\text{Conv}_{1\times1}(X, C \to 3C)),
\end{equation}
where $\text{Conv}_{1\times1}(X, \text{in}\to \text{out})$ denotes a $1\times1$ convolution layer mapping from $\text{in}$ channels to $\text{out}$ channels, and $\text{Split()}$ splits the feature map into three equal parts along the channel dimension to obtain QKV matrices.

\begin{figure*}[t]
\centering
    \centerline{\includegraphics[width=2\columnwidth]{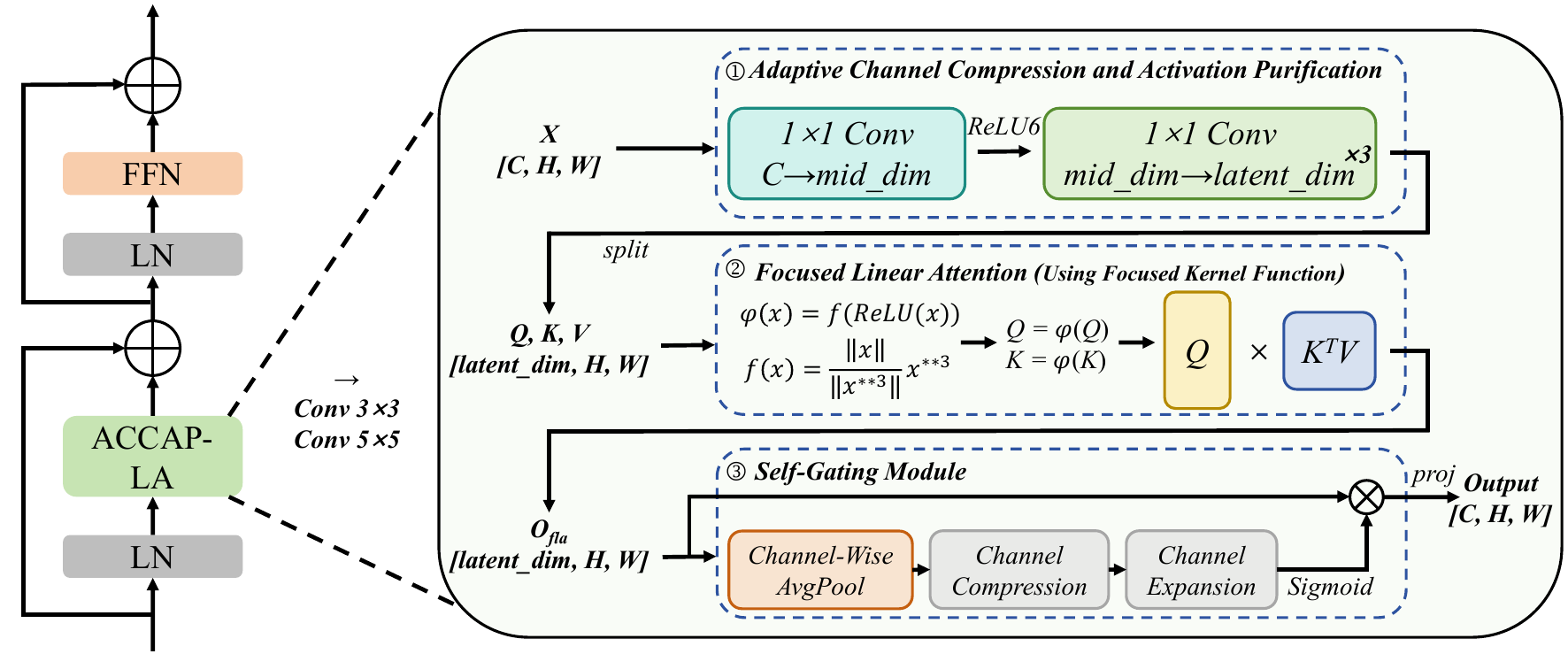}}
    \caption{
    Schematic diagram of the overall structure of our proposed attention mechanism (ACCAP-LA). The input features first undergo \(3\times3\) and \(5\times5\) convolutions to extract multi-scale features, which are then sequentially processed by three core modules: Adaptive Channel Compression and Activation Purification, Focused Linear Attention, and Self-Gating Module, to generate the final output.}
    \label{fig:ACCAP_structure}
\end{figure*}

In contrast, we integrate the ACCAP mechanism into the QKV projection process (as shown in Figure \ref{fig:ACCAP_structure}), which consists of three sequential operations.

\textbf{Step 1: Channel Compression.} A $1\times1$ convolution compresses the input channel dimension from $C$ to $\text{mid\_dim}$, where $\text{mid\_dim}=C//4$ in our experiments:
\begin{equation}
X_{\text{mid}} = \text{Conv}_{1\times1}(X, C \to \text{mid\_dim}).
\end{equation}

\textbf{Step 2:Activation Purification.} The compressed feature map is activated by the $\text{ReLU6}$ function to purify invalid activations and enhance non-linear representation:
\begin{equation}
X_{\text{act}} = \text{ReLU6}(X_{\text{mid}}) = \min(\max(X_{\text{mid}}, 0), 6).
\end{equation}

\textbf{Step 3: Channel Expansion and QKV Splitting.} A second $1\times1$ convolution excites the channel dimension from $\text{mid\_dim}$ to $\text{latent\_dim}\times3$, where $\text{latent\_dim}=C//2$, the output is then split into three equal parts to obtain QKV matrices:
\begin{equation}
X_{\text{latent}} = \text{Conv}_{1\times1}(X_{\text{act}}, \text{mid\_dim} \to 3\cdot\text{latent\_dim}),
\end{equation}
\begin{equation}
[Q, K, V] = \text{Split}(X_{\text{latent}}),
\end{equation}
where each of $Q,K,V$ has a channel dimension of $C//2$.
\subsubsection{Efficiency Gains from ACCAP}
We quantify the parameter and FLOPs savings of the ACCAP mechanism by comparing the full attention module (encompassing QKV projection and core attention computation) against standard linear attention. In alignment with mainstream ViT efficiency analysis conventions, we ignore bias terms and negligible element-wise operations in all derivations, and count each multiply-add operation as one FLOP for all computational cost statistics throughout this work.

For an input feature $X\in\mathbb{R}^{b\times n\times c}$ ($b$ = batch size, $n$ = flattened spatial sequence length, $c$ = input channels), standard linear attention uses a single 1×1 convolution to project the input to $3c$ channels for equal split into Q/K/V (each with $c$ channels), yielding $3c^2$ parameters and $3bnc^2$ FLOPs; the subsequent linear attention computation $\text{Attention}(Q,K,V) = Q\cdot (K^\mathrm{T}\cdot V)$ adds $2bnc^2$ FLOPs, resulting in a total of $3c^2$ parameters and $5bnc^2$ FLOPs for the complete module. Our ACCAP-LA replaces the single heavy projection with a lightweight two-stage 1×1 convolution pipeline (channel compression to $c/4$ followed by expansion to $3c/2$), which reduces the channel dimension of the projected Q/K/V to $c/2$ while preserving the linear attention computation paradigm. This optimized projection only introduces $5c^2/8$ parameters and $5bnc^2/8$ FLOPs, and the reduced channel dimensions cut the core attention FLOPs to $bnc^2/2$.

In total, ACCAP-LA achieves $5c^2/8$ parameters and $9bnc^2/8$ FLOPs, translating to a \textbf{79.17\%} parameter reduction and a \textbf{77.5\% }FLOPs reduction over standard linear attention. This significant efficiency gain stems from the reduced channel dimensions in both projection and attention computation, while our adaptive channel compression and activation purification strategy maintains sufficient representational capacity.

\subsubsection{Distinctions from Inverted Bottlenecks and Low-rank Projections}
To explicitly clarify the novelty boundary of ACCAP against the aforementioned channel compression paradigms, we compare it with the two most relevant structures—inverted residual bottlenecks and low-rank QKV factorization—in terms of design motivation, structural topology, and functional objectives. In contrast to inverted residual bottlenecks, which follow a \textbf{``narrow-wide-narrow''} dimension trajectory that temporarily expands channels for spatial convolution before projecting back to a low-dimensional feature output, ACCAP adopts a \textbf{``wide-narrow-wide''} channel transformation specifically engineered for the QKV projection stage: it compresses full-dimensional input features into a compact latent space for efficient computation, then expands to generate high-dimensional QKV matrices instead of returning to a low-dimensional feature map. Unlike inverted bottlenecks that act as general convolutional feature extractors in backbone stages, ACCAP is purpose-built for linear attention's QKV generation, with the core objective of mitigating linear attention's inherent representational degradation rather than solely reducing computational overhead.

Furthermore, standard low-rank QKV factorization approximates the full projection matrix via two successive linear mappings purely for parameter and computation reduction, with no non-linear feature refinement between layers. ACCAP, by contrast, inserts a ReLU6-based activation purification step between the compression and expansion projections, which filters invalid activations and refines feature semantics in the compressed latent space. Most importantly, ACCAP is not merely a low-rank approximation of the original QKV projection; it is a feature enhancement pipeline that preserves high-dimensional input semantics, reduces channels for efficiency, purifies critical feature information, and ultimately produces more expressive QKV matrices. This design directly targets the low-rank representational limitation of linear attention, rather than functioning simply as a parameter compression technique, which is also validated by the nearly unchanged feature rank across the pipeline in Figure \ref{fig:visualization}.
\subsection{Focused Linear Attention}
Vanilla linear attention abandons the softmax normalization function to achieve linear complexity, which inevitably leads to weak focus capability due to the lack of explicit similarity-based normalization between query and key vectors. 
To address this limitation and resolve the semantic ambiguity of features in low-dimensional space after ACCAP channel compression, we draw inspiration from Flatten Transformer \cite{han2023flatten} and introduce a focused kernel function $\phi(x)$ that collaborates with the ACCAP mechanism to enhance the focus ability of linear attention while preserving its $O(N)$ computational complexity. The kernel function is defined as:
\begin{equation}
    \phi(x) = \frac{\|x\|}{\|x^{**3}\|} \, x^{**3}, \label{eq:kernel_function}
\end{equation}
where $\|\cdot\|$ denotes the L2-norm, $x^{**3}$ denotes element-wise cubing. This kernel function modulates the direction of input vectors without altering their magnitude: it narrows the distance between semantically similar vectors and enlarges the gap between dissimilar ones, thereby sharpening the attention distribution and significantly boosting the focus capability of linear attention.

Based on the QKV matrices generated by the ACCAP mechanism (denoted as $Q,K,V \in \mathbb{R}^{B \times \frac{C}{2} \times H \times W}$, where $B$ represents the batch size), we integrate the kernel function $\phi(x)$ to implement focused linear attention, with the complete computation process formalized as follows:
\begin{equation}
    Q' = \phi(Q), \quad K' = \phi(K), \label{eq:focused_la1}
\end{equation}
\begin{equation}
    O_{\text{fla}} = Q'((K')^{\mathrm{T}}V). \label{eq:focused_la2}
\end{equation}

\begin{figure*}[ht]
\centering
    \centerline{\includegraphics[width=2\columnwidth]{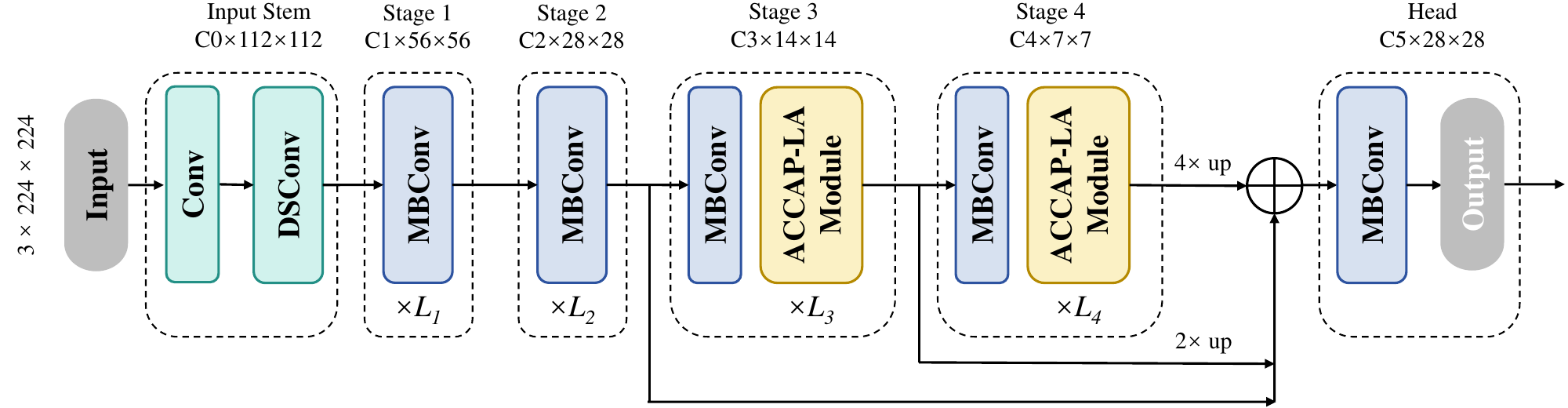}}
    \caption{
      Macro Architecture of ACCAP-ViT. $L_{1}$ to $L_{4}$ correspond to the number of repetitions of the respective modules in each stage. The ACCAP-LA module is inserted into Stages 3 and 4 of the backbone. 2x and 4x upsampling operations are applied in Stage 3 and Stage 4 for dimensionality matching.}
    \label{fig:network_architecture}
\end{figure*}
\subsection{Self-Gating Module}
Self-gating mechanisms enhance attention feature non-linearity, introduce input-dependent sparsity, and mitigate the attention sink effect while stabilizing training \cite{qiu2026gated}. To further improve efficiency and modulation precision, we integrate a channel compression-activation pipeline into the standard self-gating design:
\begin{equation}
    \mu = \frac{1}{H \times W} \sum_{i=1}^{H} \sum_{j=1}^{W} O_{\text{fla}}(C,i, j),
\end{equation}
\begin{equation}
    w = \text{Sigmoid}(W_2 \cdot \text{Sigmoid}(W_1 \cdot \mu)),
\end{equation}
\begin{equation}
    \text{Output} = O_{\text{fla}} \odot w.
\end{equation}

Specifically, we first perform global average pooling on the attention feature map $O_{\text{fla}}$ along the channel dimension, then compress the channel dimension to $\frac{1}{4}$ of its original size via linear transformation ($W_1$), apply sigmoid activation, reconstruct the channel dimension back to the original ($W_2$), and adopt sigmoid again to generate gating weights, which are finally multiplied by $O_{\text{fla}}$ to get the final output.

\subsection{Network Architecture}
We propose the ACCAP-ViT model family (M0 to M3) to validate the effectiveness of our ACCAP-LA module. All models share a unified backbone (Figure \ref{fig:network_architecture}), comprising an input stem, four hierarchical stages, and a task-adaptive head. Following the progressive feature learning paradigm, the backbone downsamples feature maps and expands channel dimensions as the network depth increases.

Stage 1 and Stage 2 are composed of MBConv blocks, which extract low-level visual features (e.g., edges, textures and basic shapes) and perform initial spatial downsampling with progressive channel expansion. This lays the necessary foundation for subsequent high-level semantic feature learning.
Stage 3 and Stage 4 take our proposed ACCAP-LA module as the core component, with each module equipped with a residual connection. The ACCAP-LA module is designed to model complex long-range spatial dependencies and realize adaptive multi-scale semantic feature fusion, which is critical for high-level visual understanding.

To facilitate effective multi-scale feature fusion, the outputs of Stage 3 and Stage 4 are upsampled by factors of 2 and 4, respectively, to achieve dimension alignment. In this work we adopt an image classification head for training. The backbone can be easily extended to diverse downstream tasks by replacing this head with task-specific ones.

The detailed configurations of ACCAP-ViT-M0 to M3 are summarized in Table \ref{tab:arch_config}, where we define the key structural hyperparameters for the model family. Specifically, $ C_0 $ to $ C_4 $ denote the channel dimensions of the corresponding stages in the backbone, $ L_1 $ to $ L_4 $ represent the number of stacked blocks in Stage 1 to Stage 4, $ H_1 $ and $ H_2 $ represent the hidden dimension settings in the classification head.

\begin{table}[h]
\caption{Architecture details of ACCAP-ViT model variants.}
\centering
\scalebox{0.75}{\begin{tabular}{cccc}
  \hline
  Model & $\{C_0,C_1,C_2,C_3,C_4\}$ & $\{L_1,L_2,L_3,L_4\}$ & $\{H_1,H_2\}$ \\
  \hline
  M0 & $\{16,32,48,64,128\}$ & $\{2,3,3,3\}$ & $\{512,1024\}$ \\
  M1 & $\{16,32,64,128,192\}$ & $\{2,3,6,2\}$ & $\{768,1152\}$ \\
  M2 & $\{16,32,64,128,192\}$ & $\{3,3,10,4\}$ & $\{768,1152\}$ \\
  M3 & $\{32,64,128,192,256\}$ & $\{2,3,10,3\}$ & $\{768,1152\}$ \\
  \hline
\end{tabular}}
\label{tab:arch_config}
\end{table}

\begin{table}[ht]
\centering
\caption{Summary of experimental configurations and hyperparameters}
\label{tab:exp_config}
\scalebox{0.8}{\begin{tabular}{llc}
\hline
Category & Item & Setting \\
\hline
\multirow{3}{*}{Computing Env.}
& Software & PyTorch 2.0.1, CUDA 12.2 \\
& CPU & Intel Xeon Platinum 8280 \\
& GPU & 8$\times$ RTX 3090 \\
\hline
\multirow{1}{*}{Eval. Protocol}
& Resolution & 224$\times$224 \\
\hline
\multirow{6}{*}{ImageNet-1K}
& Optimizer & AdamW \\
& Epochs & 300 \\
& Batch Size & 128 \\
& EMA Decay & 0.9998 \\
& Label Smooth & 0.1 \\
& DropPath Rate & 0.05 (linear decay) \\
\hline
\multirow{4}{*}{Architectural Configs}
& Head Dim. & 16 \\
& MLP Ratio & 4 \\
& MBConv Ratio & 4 \\
& MBConv Act. & h-swish \\
\hline
\multirow{3}{*}{COCO 2017}
& Framework & MMDetection \\
& Schedule & RetinaNet 1$\times$ \\
& Backbone & ImageNet-1K pre-trained \\
\hline
\end{tabular}}
\end{table}
\section{Experiments}
\begin{table*}[h]
\caption{Comparison of ACCAP-ViT models with existing efficient CNNs, ViT and Mamba models on the ImageNet-1K dataset.}
\centering
\begin{tabular}{lcccc}
\hline
\textbf{Models} & \textbf{Top-1 (\%)} & \textbf{Top-5 (\%)} & \textbf{FLOPs (M)} & \textbf{Params (M)} \\
\hline
EfficientViT-M0 \cite{liu2023efficientvit} & 63.2 & 85.4 & 79 & 2.3 \\
S3ML-ViT-M0 \cite{zhang2025s3ml} & 67.1 & 87.4 & 110 & 2.5 \\
Mobile-Former-52M \cite{chen2022mobile} & 68.7 & - & 52 & 3.5 \\
ShuffleNetV2 1.0x \cite{ma2018shufflenet} & 69.4 & 88.9 & 146 & 2.3 \\
MobileViTV2-0.5 \cite{mehta2022separable} & 70.2 & - & 466 & 1.4 \\
\textbf{ACCAP-ViT-M0 (ours)} & \textbf{72.3} & \textbf{90.7} & 191 & 2.4 \\
\hline
S3ML-ViT-M1 \cite{zhang2025s3ml} & 71.6 & 90.3 & 190 & 4.4 \\
MobileNetV2 1.0x \cite{sandler2018mobilenetv2} & 72.0 & 91.0 & 300 & 3.4 \\
HLSM-ViT-M1 \cite{Guan2026HybridLA} & 72.5 & 90.8 & 186 & 4.3 \\
StarLA-ViT-S1 \cite{guan2026star} & 72.7 & 91.0 & 185 & 4.3 \\
GhostNet 1.0x \cite{han2020ghostnet} & 73.9 & 91.4 & 141 & 5.2 \\
MicroViT-S2 \cite{setyawan2025microvit} & 74.6 & - & 345 & 10.0 \\
InceptionNeXt-A \cite{DBLP:conf/cvpr/YuZYW24} & 75.3 & - & - & 4.2 \\
\textbf{ACCAP-ViT-M1 (ours)} & \textbf{77.5} & \textbf{93.7} & 432 & 4.3 \\
\hline
EfficientViT-M4 \cite{liu2023efficientvit} & 74.3 & 91.8 & 299 & 8.8 \\
ShuffleNetV2 2.0x \cite{ma2018shufflenet} & 74.9 & 92.4 & 591 & 7.4 \\
VCA-DeiT-T \cite{pu2025lineardiff} & 75.6 & - & 1200 & 6.0 \\
S3ML-ViT-M2 \cite{zhang2025s3ml} & 76.6 & 93.2 & 480 & 6.0 \\
StarNet-S3 \cite{ma2024rewrite} & 77.3 & - & 757 & 5.8 \\
StarLA-ViT-S2 \cite{guan2026star} & 77.7 & 93.8 & 479 & 5.8 \\
\textbf{ACCAP-ViT-M2 (ours)} & \textbf{79.4} & \textbf{94.8} & 637 & 5.8 \\
\hline
SHViT-S3 \cite{yun2024shvit} & 77.4 & 93.4 & 601 & 14.2 \\
FLatten-PVT-T \cite{han2023flatten} & 77.8 & 93.9 & 2000 & 12.2 \\
Agent-PVT-T \cite{han2024agent} & 78.4 & - & 2000 & 11.6 \\
MobileMamba-S6 \cite{he2025mobilemamba} & 78.0 & - & 652 & 15.0 \\
StarLA-ViT-S3 \cite{guan2026star} & 79.5 & 94.5 & 691 & 8.7 \\
HLSM-ViT-M3 \cite{Guan2026HybridLA} & 79.5 & 94.6 & 694 & 8.6 \\
VRWKV-S \cite{DBLP:conf/iclr/DuanWCZLLQ0DW25} & 80.1 & - & 4600 & 23.8 \\
VCA-DeiT-S \cite{pu2025lineardiff} & 80.7 & - & 4600 & 22.6 \\
PACA-Tiny \cite{Grainger2023PaCaViT} & 80.9 & - & 3200 & 12.2 \\
\textbf{ACCAP-ViT-M3 (ours)} & \textbf{81.4} & \textbf{95.7} & 1490 & 9.0 \\
\hline
\end{tabular}
\label{tab:master-table}
\end{table*}

In this section, we conduct a comprehensive suite of experiments to systematically validate the effectiveness, transfer learning capability, and cross-architecture compatibility of the proposed ACCAP-LA module. All implementations are built on PyTorch 2.0.1 with CUDA 12.2, and all training, validation, and testing procedures are executed on a server equipped with an Intel Xeon Platinum 8280 CPU and 8 NVIDIA RTX 3090 GPUs. Parameter counts and FLOPs are computed under a unified 224×224 input resolution following standard conventions in lightweight vision models. Core implementation configurations and training hyperparameters are summarized in Table \ref{tab:exp_config} for complete reproducibility.

We first instantiate the ACCAP-ViT model family and perform pre-training on the ImageNet-1K dataset to verify its performance on the fundamental image classification task. We then transfer the ImageNet-1K pre-trained ACCAP-ViT backbones to the object detection task for comparative experiments with existing efficient models, evaluating the module's utility on downstream tasks. 

Furthermore, we carry out extensive ablation studies to dissect the key design choices and component contributions of ACCAP-LA. Specifically, we first ablate individual modules of ACCAP-LA, to quantify the performance gain brought by each component. On this basis, we further investigate the impact of critical hyperparameter configurations (e.g., different compression and Expansion ratios) on performance and integrate the ACCAP-LA module into classic lightweight vision transformers (PVT \cite{wang2022pvt}, EfficientViT \cite{cai2023efficientvit}, Swin \cite{liu2021swin}) to validate its plug-and-play generalizability. The detailed experimental settings, results, and in-depth analysis for each of the above experiments are presented in the following subsections.

\subsection{ImageNet-1K Classification}
We conduct the image classification experiment on the ImageNet-1K dataset, which consists of 1.2 million training images and 50K validation images across 1000 object categories. To fully evaluate the performance of our ACCAP-ViT, we compare it with existing efficient models based on CNNs, ViT, Mamba, and their hybrid architectures under the same experimental setup. Specifically, we adopt the AdamW optimizer for training; the training process running for 300 epochs, with a batch size of 128 and the input image resolution uniformly resized to 224×224. The comprehensive comparison results in terms of Top-1/Top-5 accuracy, Floating-point Operations (FLOPs, M), and number of parameters (Params, M) are presented in Table~\ref{tab:master-table}.

\begin{table*}[htb]
\caption{Comparison of ACCAP-ViT Object Detection Performance on COCO 2017 with Existing Efficient Models.}
\centering
\begin{tabular}{lcccccccc}
\hline
\textbf{Models} & \multicolumn{6}{c}{\textbf{RetinaNet 1×}} & \textbf{FLOPs} & \textbf{Params} \\
 & \textbf{AP} & \textbf{AP$_{50}$} & \textbf{AP$_{75}$} & \textbf{AP$_s$} & \textbf{AP$_m$} & \textbf{AP$_l$} & \textbf{(M)} & \textbf{(M)} \\
\hline
MobileNetV3 \cite{howard2019searching} & 29.9 & 49.3 & 30.8 & 14.9 & 33.3 & 41.1 & 217 & 5.4 \\ 
MNASNet-A2 \cite{tan2019mnasnet} & 30.5 & 50.2 & 32.0 & 15.6 & 34.1 & 41.0 & 340 & 4.8 \\
SPOS \cite{guo2020single} & 30.7 & 49.8 & 32.2 & 15.4 & 33.9 & 41.6 & 365 & 4.3 \\
FairNAS-C \cite{chu2021fairnas} & 31.2 & 50.8 & 32.7 & 16.3 & 34.4 & 42.3 & 325 & 5.6 \\
S3ML-ViT-M1 \cite{zhang2025s3ml} & 31.5 & 49.6 & 33.2 & 17.2 & 33.9 & 42.2 & 190 & 4.4\\
EfficientViT-m4 \cite{liu2023efficientvit} & 32.7 & 52.2 & 34.1 & 17.6 & 35.3 & 46.0 & 299 & 8.8 \\
HLSM-ViT-M1 \cite{Guan2026HybridLA} & 33.2 & 52.6 & 35.1 & 18.0 & 36.2 & 44.4 & 186 & 4.3 \\
StarLA-ViT-S1 \cite{guan2026star} & 33.3 & 52.8 & 34.7 & 19.0 & 36.4 & 44.2 & 186 & 4.3 \\
\hline
\textbf{ACCAP-ViT-M1 (ours)} & \textbf{36.0} & \textbf{55.5} & \textbf{38.1} & \textbf{20.2} & \textbf{39.9} & \textbf{46.7} & 432 & 4.3 \\
\hline
\end{tabular}
\label{tab:object-detection}
\end{table*}

Across all model scales (M0 to M3), our ACCAP-ViT consistently achieves the best Top-1 and Top-5 accuracy among counterparts with comparable parameter counts or FLOPs, demonstrating a superior performance-efficiency trade-off on ImageNet-1K. For lightweight models, ACCAP-ViT-M0 outperforms EfficientViT-M0 \cite{liu2023efficientvit} and S3ML-ViT-M0 \cite{zhang2025s3ml} by 9.1\% and 5.2\% in Top-1 accuracy respectively. For medium and large scales (M1-M3), our models maintain this leading trend: ACCAP-ViT-M1 (4.3M params) achieves a 77.5\% Top-1 accuracy, outperforming InceptionNeXt-A \cite{DBLP:conf/cvpr/YuZYW24} by 2.2\%; ACCAP-ViT-M3 even hits an 81.4\% Top-1 and 95.7\% Top-5 accuracy with only 9.0M params, outperforming large models like VRWKV-S (23.8M params, 80.1\% Top-1) \cite{DBLP:conf/iclr/DuanWCZLLQ0DW25} and Vim-S (26.0M params, 80.3\% Top-1) \cite{DBLP:conf/icml/ZhuL0W0W24} with much fewer parameters and lower FLOPs.

Notably, our ACCAP-ViT outperforms not only vanilla ViT and CNN-based models but also recent Mamba \cite{he2025mobilemamba} and hybrid architectures \cite{zhang2025s3ml}, which validates the effectiveness of the proposed ACCAP-LA mechanism. Unlike existing models that often sacrifice accuracy for efficiency, ACCAP-ViT realizes a simultaneous improvement in both metrics, making it more suitable for resource-constrained visual classification tasks.

\subsection{COCO 2017 Object Detection}
We perform object detection experiments using the RetinaNet framework, with our ACCAP-ViT adopted as the backbone network. The experiments are conducted on the COCO 2017 dataset, which comprises 118K training images and 5K validation images. To facilitate effective training and transfer learning, our ACCAP-ViT-M1 backbone is initialized with pre-trained weights on ImageNet-1K. Following previous work \cite{yu2022metaformer}, we employ the AdamW optimizer with an initial learning rate of 0.0002, and the model is trained for 12 epochs. We compare ACCAP-ViT-M1 with representative efficient CNN and ViT-based backbones under the same RetinaNet (1× schedule) setup. Detailed results are summarized in Table~\ref{tab:object-detection}.

With a slightly lower parameter count (4.3 M) compared to S3ML-ViT-M1 (4.4 M) \cite{zhang2025s3ml}, our ACCAP-ViT-M1 achieves comprehensively superior detection performance. It attains an overall AP of 36.0\%, which outperforms all compared models regardless of architecture type. All associated detection metrics of ACCAP-ViT-M1 also rank first, confirming its excellent capability in feature representation for object detection tasks.

ACCAP-ViT-M1 does not perform well in terms of FLOPs (432M), and this is attributed to the number of stacked blocks designed for each stage during the backbone development. In lightweight efficient network design, parameter count is often a more critical consideration. This is because parameters directly correlate with model memory footprint, deployment difficulty and inference latency, which are core constraints in real-world applications. We thus specifically select models with similar parameter counts for fair comparison. Its strong performance on object detection tasks further proves that it possesses great potential for migration to downstream visual tasks.
\subsection{Ablation Study}
\subsubsection{Ablation Study on Core Modules}
To quantify the contribution of each core component in ACCAP-ViT and clarify their individual effects on model performance, we conduct ablation experiments on the ImageNet-1K dataset based on ACCAP-ViT-M1. Specifically, we perform single-module removal experiments: each time only one key module is removed while all other components remain intact. The ablated modules include ACCAP module, the Focused Kernel Function, Multi-Scale Feature module, and Self-Gating module. We also ablate the activation function of the Self-Gating module and replace the original ReLU6() (used in ACCAP) with tan() to explore the impact of activation function selection within ACCAP. The results, presented as Top-1 accuracy degradation relative to the baseline, are summarized in Table~\ref{tab:ablation_study}.

\begin{table}[h]
\caption{Ablation Study on ImageNet-1K: Top-1 Accuracy Comparison for Each Module Removal.}
\centering
\begin{tabular}{lcc}
\hline
\textbf{Module Removed} & \textbf{Top-1 (\%)} \\
\hline
\textbf{ACCAP-ViT-M1} & \textbf{77.488} \\
w/o ACCAP & -3.838 \\
w/o Focused Kernel Function & -0.258 \\
w/o Multi-Scale Feature & -0.692 \\
w/o Self-Gating & -0.562 \\
w/o Self-Gating's activation function & -0.284\\
ACCAP's ReLU6() → tan() & -0.280 \\
\hline
\end{tabular}
\label{tab:ablation_study}
\end{table}

As shown in Table~\ref{tab:ablation_study}, removing different modules causes varying performance degradation, and the ACCAP exerts the most prominent influence (-3.838\%). The Multi-Scale Feature module contributes notably (0.692\% reduction), validating the value of multi-scale input features. Removing the Self-Gating module results in a 0.562\% drop, highlighting its role in adaptive feature adjustment. The Focused Kernel Function, the Self-Gating activation function and the replacement of ReLU6() to tan() have minor impacts, serving as auxiliary fine-tuning components. These results confirm the rationality of each module design, with ACCAP being the pivotal driver of the model's performance advantage.

\subsubsection{Generalization Verification of ACCAP-LA}
To validate the plug-and-play capability and generalizability of the proposed ACCAP-LA mechanism, we replace the original attention modules of three classic lightweight vision transformers---EfficientViT-b1 \cite{cai2023efficientvit}, PVT-T \cite{wang2021pyramid}, and Swin-T \cite{liu2021swin}---with ACCAP-LA, and evaluate their performance on ImageNet-1K (the models with ``*'' denote the variants after replacement). 

\begin{table}[h]
\caption{Generalization Verification of ACCAP-LA by Replacing Attention Mechanisms of classic methods.}
\centering
\begin{tabular}{lcc}
\hline
\textbf{Models} & \textbf{Top-1 Accuracy (\%)} & Params (M) \\
\hline
EfficientViT-b1 & 78.1 & 9.1 \\
EfficientViT-b1* & \textbf{81.4(+3.3)} & 9.0 \\
PVT-T & 75.1 & 13.2 \\
PVT-T* & \textbf{78.5(+3.4)} & 13.2\\
Swin-T & 80.8 & 28.3 \\
Swin-T* & \textbf{81.6(+0.8)} & 29.1 \\
\hline
\end{tabular}
\label{tab:pvt}
\end{table}

As shown in Table~\ref{tab:pvt}, ACCAP-LA brings consistent performance improvements across all tested models, demonstrating its effectiveness and compatibility with different transformer architectures. Specifically, the performance gains are more significant in models with smaller parameter counts: EfficientViT-b1* achieves a Top-1 accuracy of 81.4\% (+3.3\%) with slightly reduced parameters (9.0M vs. 9.1M), and PVT-T* obtains an accuracy of 78.5\% (+3.4\%) while maintaining the same parameter count. In contrast, Swin-T* only gains 0.8\% accuracy, indicating that ACCAP-LA is currently more suitable for small-scale models. How to further extend the effectiveness of ACCAP-LA to large-scale models with more complex structures remains a valuable direction for future research.

\subsubsection{Computational Efficiency Comparison}
To quantify the computational efficiency of ACCAP-LA, we compare its inference speed against six mainstream attention mechanisms \cite{liu2023efficientvit,Guan2026HybridLA,zhang2025s3ml,wang2022pvt,han2024agent,liu2021swin} under a unified setting. For fairness, we test only the attention module with identical input (batch size 32, 128 channels, 56×56 feature map) on a single NVIDIA RTX 4090 GPU, reporting per-sample latency and throughput.

\begin{table}[h]
\caption{Computational Efficiency Comparison of Different Attention Modules.}
\centering
\scalebox{0.8}{\begin{tabular}{lcc}
\hline
\textbf{Method} & Latency↓ (ms) & Throughput↑ (samples/s) \\
\hline
EfficientViT & 0.8222 & 1216.23 \\
HLSM-ViT & 0.4067 & 2458.70 \\
S3ML-ViT & 0.3780 & 2645.40 \\
PVT & 0.1637 & 6107.67 \\
Agent-PVT & 0.1244 & 8038.12 \\
Swin & 0.0991 & 10092.15 \\
\textbf{ACCAP-LA (Ours)} & \textbf{0.0788} & \textbf{12683.38} \\
\hline
\end{tabular}}
\label{tab:efficiency}
\end{table}

As shown in Table \ref{tab:efficiency}, ACCAP-LA achieves the lowest latency (0.0788 ms/sample) and highest throughput (12683.38 samples/s), outperforming all baselines by a clear margin and validating its superior computational efficiency.

\subsubsection{Ablation Study on Compression and Expansion Ratios}
To optimize the hyperparameter configuration of ACCAP and clarify the impact of channel compression and expansion ratios on model performance, we conduct ablation experiments on these two key ratios. We set multiple ratio combinations for compression and expansion respectively, with ACCAP-ViT-M1 as the baseline. Table~\ref{tab:sa} presents the variation of Top-1 accuracy relative to the baseline (77.488\%), where negative values indicate performance degradation.
\begin{table}[h]
\caption{Ablation Study on Compression and Expansion Ratios in ACCAP.}
\centering
\scalebox{0.8}{\begin{tabular}{cccc}
\hline
\textbf{Compression Ratio\textbackslash Expansion Ratio} & \textbf{1/2} & \textbf{1/4} & \textbf{1/8} \\
\hline
\textbf{1/2} & -1.664 & -0.170 & -0.500 \\
\textbf{1/4} & 77.488 & -0.476 & -0.594 \\
\textbf{1/8} & -0.198 & -0.804 & -0.936 \\
\textbf{1/16} & -0.630 & -0.868 & -0.910 \\
\hline
\end{tabular}}
\label{tab:sa}
\end{table}

As indicated in Table~\ref{tab:sa}, the compression and expansion ratios significantly affect the performance of the model, and overall performance tends to degrade as both ratios decrease. The combination of 1/4 compression ratio and 1/2 expansion ratio achieves the best performance (consistent with the baseline), confirming it as the optimal configuration for ACCAP. In particular, the 1/2 compression + 1/2 expansion combination suffers a substantial accuracy drop of 1.664\%, which is attributed to the cancellation of the subsequent expansion process in this group. All other combinations retain the complete ACCAP pipeline with only ratio adjustments. This result demonstrates that mere channel compression cannot ensure good performance and that the activation purification component in ACCAP plays a critical role in improving the quality of the features.
\begin{figure}[t]
    \centering
    \centerline{\includegraphics[width=1\columnwidth]{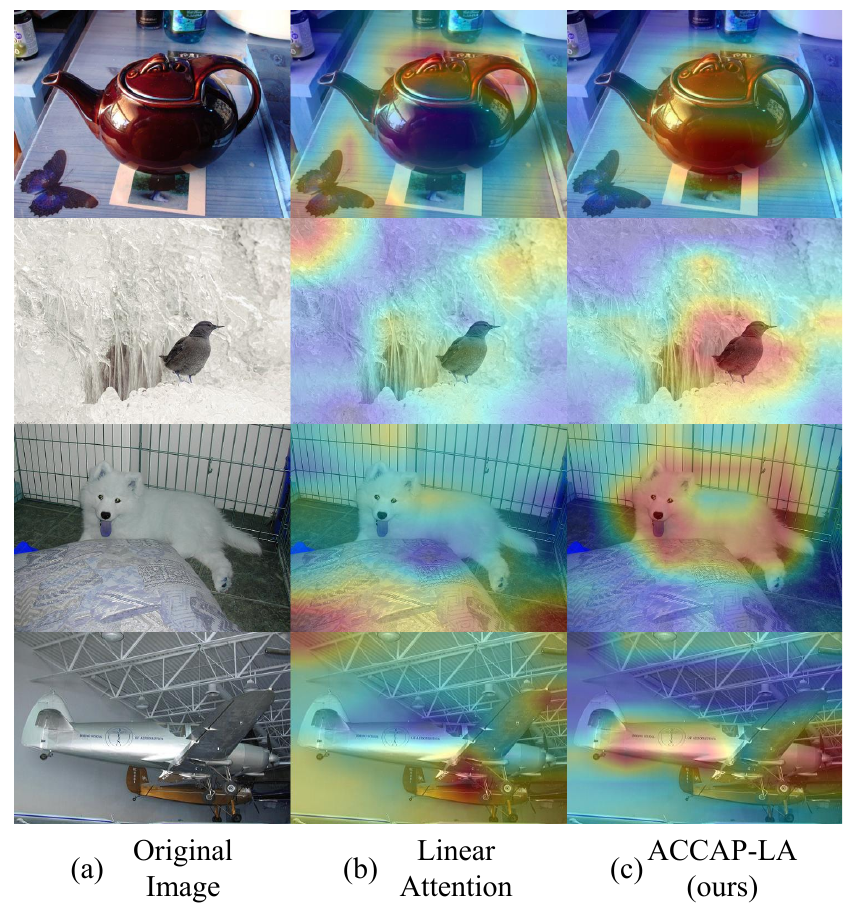}}
    \caption{Visualization comparison of attention feature maps. Columns from left to right: (a) original image, (b) vanilla linear attention, (c) our proposed ACCAP-LA.}
    \label{fig:attention_visualization}
\end{figure}
\subsection{Attention Visualization}
To provide supplementary qualitative verification beyond our quantitative ablation study, we conduct visualization analysis of attention feature maps using samples from the ImageNet-1K validation set, comparing vanilla linear attention and our proposed ACCAP-LA. As shown in Figure~\ref{fig:attention_visualization}, our ACCAP-LA empowered by the focused kernel function achieves significantly enhanced focusing capability, which can accurately focus on key target regions in images. Meanwhile, the ACCAP mechanism effectively filters redundant feature information and suppresses the model's attention to non-target background areas. These qualitative observations are consistent with our quantitative experimental results, further validating the effectiveness of our method.

\section{Conclusion}
To address the efficiency-capability dilemma of attention mechanism in Vision Transformers, this paper proposes ACCAP-LA, a novel lightweight linear attention mechanism that for the first time embeds the Adaptive Channel Compression and Activation Purification (ACCAP) mechanism into the QKV projection process. This design fundamentally mitigates the representational degradation of vanilla linear attention while achieving substantial reductions in computational overhead and parameter count. Extensive experiments across image classification and object detection tasks validate that the proposed ACCAP-ViT series delivers superior efficiency-capability trade-offs compared to state-of-the-art CNNs, ViT, and Mamba-based efficient models. Comprehensive ablation studies confirm the rationality of each component's design and the crucial role of ACCAP in driving performance gains, while cross-architecture generalization experiments demonstrate ACCAP-LA's plug-and-play compatibility with classic lightweight transformers. Future work will delve deeper into the underlying mechanisms of adaptive channel compression and activation purification, and extend the proposed framework to large-scale models, aiming to further advance the development and practical application of efficient Vision Transformers.

{
\bibliographystyle{unsrt}
\bibliography{egbib}
}

\end{document}